\documentclass[letterpaper, 10 pt, conference]{ieeeconf}  % Comment this line out if you need a4paper

\IEEEoverridecommandlockouts                      

\usepackage{graphics} % for pdf, bitmapped graphics files
\usepackage{epsfig} % for postscript graphics files
\usepackage{mathptmx} % assumes new font selection scheme installed
\usepackage{times} % assumes new font selection scheme installed
\usepackage{amsmath} % assumes amsmath package installed
\usepackage{amssymb}  % assumes amsmath package installed
\usepackage{mathrsfs}
\newcommand{\p}{\pmb}
\usepackage{multirow}
\usepackage{algorithm, algorithmic}
\usepackage{subfig}

\title{\LARGE \bf
A Statistical Method for Parking Spaces Occupancy Detection via Automotive Radars
}

\author{Qi Luo$^{1}$, \emph{Student Member, IEEE}, Romesh Saigal$^{1}$, Robert Hampshire$^{2}$ and Xinyi Wu $^{1}$% <-this % stops a space
\thanks{*This work was supported by Mobility Transformation Center, University of Michigan.}% <-this % stops a space
\thanks{$^{1}$ Qi Luo, Romesh Saigal and Xinyi Wu are with the Department of Industrial and Operations Engineering,
        University of Michigan, Ann Arbor, MI 48109.
        Email: {\tt\small rsaigal@umich.edu}}%
\thanks{$^{2}$ Robert Hampshire is with Transportation Research Institute, University of Michigan,
        Ann Arbor, MI 48109.
        Email: {\tt\small hamp@umich.edu}}%
}

\begin{document}

\maketitle
\thispagestyle{empty}
\pagestyle{empty}

%%%%%%%%%%%%%%%%%%%%%%%%%%%%%%%%%%%%%%%%%%%%%%%%%%%%%%%%%%%%%%%%%%%%%%%%%%%%%%%%
\begin{abstract}
Real-time parking occupancy information is valuable for guiding drivers' searching for parking spaces. Recently many parking detection systems using range-based on-vehicle sensors are invented, but they disregard the practical difficulty of obtaining access to raw sensory data which are required for any feature-based algorithm. In this paper, we focus on a system using short-range radars (SRR) embedded in Advanced Driver Assistance System (ADAS) to collect occupancy information, and broadcast it through a connected vehicle network. The challenge that the data transmitted through ADAS unit has been encoded to sparse points is overcome by a statistical method instead of feature extractions. We propose a two-step classification algorithm combining Mean-Shift clustering and Support Vector Machine to analyze SRR-GPS data, and evaluate it through field experiments. The results show that the average Type I error rate for off-street parking is $15.23 \%$ and for on-street parking is $32.62\%$. In both cased the Type II error rates are less than $20 \%$. Bayesian updating can recursively improve the mapping results. This paper can provide a comprehensive method to elevate automotive sensors for the parking detection function.     
\newline    
\end{abstract}


%%%%%%%%%%%%%%%%%%%%%%%%%%%%%%%%%%%%%%%%%%%%%%%%%%%%%%%%%%%%%%%%%%%%%%%%%%%%%%%%
\section{INTRODUCTION}

Traffic resulting from vehicles hunting for free parking spaces is significant in populated urban areas. Previous research reported that vehicles cruising for parking induce about 30\% of the traffic in several major cities and consequently impose unnecessary high costs to drivers and transport authorities \cite{shoup2006cruising}. Real-time parking information provides critical input for a parking management system, which was collected by on-site sensors in the past. However, installations of sensors is costly. Besides, optimizing routing for parking with low market penetration is a significant problem for autonomous vehicles (AVs), which can be improved by collaborative mapping of multiple probe cars. Recently, many researchers have investigated the infrastructure-independent parking detection system in which non-AVs can be transformed to probe cars by mounting range-based or vision-based sensors. Furthermore, these sensors have been widely embedded in Advanced Driving Assistance Systems (ADAS), while the potential of sensory data collected in the last mile of trips is not fully exploited for parking guidance services. The major obstacle is that the access to ADAS's raw data is usually not available, and thus any previous developed algorithms based on feature extractions will require installing new sensors on the non-AVs instead. To develop an universal and economic solution, we focus on employing sparse ADAS-processed output directly.  Hence this paper intends to fill this gap by evaluating a statistical parking detection algorithm by analyzing pre-processed Short Range Sensors (SRR) data collected in experiments. \par
This paper will first review related works in section II, and propose a parking guidance system hinged on multiple sensors and a two-step clustering algorithm in section III. In section IV, this algorithm is evaluated by training data and test data from field experiments. A final conclusion is given in section V.    

\section{Related Works}

Previous studies have employed different methods for parking space detection via on-vehicle sensors. Vision-based parking detection is one of the most well-developed methods because of the maturity of machine visions for object recognition, and the lower price of cameras compared to other sensors \cite{wu2006parking, ichihashi2010improvement}. However, poor lighting or weather conditions and  deformation or occlusion objects can restrict the use of a single type of sensing technology. \par 

Range-based sensors, including ultrasonic, radar and lidar sensors, are complementary solutions to this task. For example, Schmid et al. implemented three automotive-use short-range radars operating at 24 GHz to reconstruct a hierarchical 3-D occupancy grid map with dynamic level of details \cite{schmid2011parking}.  Mathur et al. collected 500 miles of road-side parking data by equipping ultrasonic sensors on probe cars and the result showed that parking spot counts are 95\% accurate and occupancy maps can achieve over 90\% accuracy \cite{mathur2010parknet}.  Zhou et al. used AdaBoost algorithm to train a classifier on 2-D laser scans, and extracted car bumpers as main features of parked vehicles   \cite{ zhou2012detection}. Thronton et al. applied laser sensor for the fast survey of parallel on-street parking. They focused on filtering out road curbs and other driving cars on street as noise \cite{thornton2014automated}. Ibisch et al. employed RANSAC and Kalman Filters in tracking parking through multiple Lidar sensors embedded in a parking garage  in the lack of GPS information  \cite{ibisch2013towards}. \par 

It is noticed that all these experiments analyzed raw sensor data rather than from commercial product output channels. Also most literature tested their method in a single type of parking, while there are few evaluations on the overall performance when driving in different environment. 

\section{Parking Detection Algorithm and Evaluation Procedures}
In this paper, we considered two types of parking: off-street parking (open parking lots) and on-street parking (road-side parkings). These two scenarios include both parallel parking and perpendicular parking. Limited by the capability of less advanced range-based sensors (such as radars or ultrasonic sensors) embedded in commercial ADAS and absence of GPS indoors, these probe cars will require additional devices to be able to navigate inside of a multi-level parking structure. On the other hand, a multi-storey parking structures can be directly integrated into a urban-level management system by sharing status of counting at tolling gate, which however is not feasible for the two scenarios we considered here.\par

The proposed parking detection system consists of three-types of sensors that are already widely equipped in passenger vehicles: (a) Range-based sensors, including radars or ultrasonic sensors  (incorporated in ADAS such as Lane-keeping / Lane-change Assist Systems, Parking Assist Systems, Autonomous Emergency Braking Systems, etc.).  (b) Odometers, such as Global Positioning System (GPS). (c) Vehicle-to-vehicle and vehicle-to-infrastructure communication channels like Dedicated Short Range Communications. Although this paper only evaluated SRR, the conclusion should be more general for any similar combination of sensors in the list. 

\subsection{Parking Detection Algorithm for Probe Car}
Different from previous parking detection projects using the raw data, the SRR data from ADAS output are sparse 2-D points accumulated in time. Each of the points represents an object in view instead of multiple reflections from the same object at each time step. Therefore, it is not suitable to apply feature-based classification algorithms as in the related works to process data from such a  ``black box''. Alternately, a statistical detection method for the probe car can be decomposed to following steps:
\begin{enumerate}
\item[] \hspace*{-0.33in} \emph{Step 1:} Data Preprocessing: converting radar data from local coordinates to a global coordinate and synchronizing time steps between different sensors. 
\item[] \hspace*{-0.38in} \emph{Step 2:} Step-One Classification: labeling data points to clusters using a certain clustering method.     
\item[] \hspace*{-0.35in} \emph{Step 3:} Step-Two Classification: using Support Vector Machine (SVM) to find the linear maximum margins between adjacent clusters.
\item[] \hspace*{-0.35in} \emph{Step 4:} Map Matching: match margins locations with a given parking spaces map. 
\end{enumerate}

\subsubsection{Coordinate Conversion}  Since SRR data are reported in a local coordinate whose origin is the centroid of the probe car, it is necessary to project them to a global 2-D Cartesian coordinate system by combining with GPS data at each time step. We use Universal Transverse Mercator (UTM) system so that all measurements are in SI units. \par

Let $(x_t, y_t)$ denote the centroid position of the probe car at time $t$ (UTM plane), and $\phi_t$ is the heading. $\p{z}_t^* = \{ z_{t, x}^*, z_{t, y}^*\}$ represents a detected object's position in the local coordinate. A simple conversion of $\p{z}_t$ in the global coordinate is:
\begin{align*}
\begin{cases}
	z_{t, x} = \| \p{z}_t^* \| \cos (\phi_t + \theta) \\
	z_{t, y} = \| \p{z}_t^* \| \sin (\phi_t + \theta) \\
\end{cases} 
\mbox{ in which }\theta = \arctan (\frac{z_{t, y}^*}{z_{t, x}^*}) .
\end{align*}

Clocks synchronization between different types of sensor is necessary because GPS (10 Hz) has lower frequency than SRR (50 Hz). We use GPS clock as the standard timer, and $\pm 0.1 $s tolerance in matching data from different sources.  
\vspace*{0.1in} 

\subsubsection{Two-step Classification} Converted SRR data points are generated from objects within the range of the probe car's sensors, which are treated as clusters. The task of classification includes two part: finding the number of clusters (i.e. parked vehicles) at the end of the trip and labeling each point to the corresponding cluster.  \par

The first step is applying clustering with certain criterion as an unsupervised classification in the observations. To find the most efficient combination for this specific case, we compared three different types of methods in training dataset: mean-shift clustering (MSC) with flat kernel (density-based clustering method), Gaussian Mixture Model (GMM) with Akaike Information Criterion (AIC) model selection (distribution-based clustering), and distance-based K-means clustering (K-means) \cite{zivkovic2004improved,cheng1995mean,comaniciu2002mean}. In addition, it is observed that GMM is very sensitive to outliers on the perpendicular direction to the probe vehicle, which is improved by collapsing perpendicular direction of in clustering.  All four methods will return both optimal number of clusters and corresponding labels for each data point. \par

We pay special attention to MSC because it is the best fit for fixed-size (size of cars) and non-ellipse clusters case. Oppositely, K-means is not able to detect non-spherical clusters. and the accuracy of distribution-based clustering like GMM depends on the proximity of prior distribution to the formulation of data. Besides GMM and K-means need to iteratively select the optimal number of clusters according to certain criterion, which can be avoided by using MSC. \par
MSC is a nonparametric feature-space clustering technique working as a mode-seeking process. The weighted mean of the density in the neighborhood $N(x_c)$ within $\lambda$-ball of point $x_c$ is:
\begin{align*}
m(x_c) = \frac{\sum_{x_i \in N(x_c)} K(x_i - x_c)x_i}{\sum_{x_i \in N(x_c)}K(x_i - x_c)}.
\end{align*}    

The difference $m(x_c) - x_c$ is called mean shift, and in each iteration $x_c \leftarrow m(x_c)$ is performed for all data points until $m(x_c)$ converges. \par

One disadvantage of MSC is low computational efficiency. Inasmuch as that, we applied the simplest characteristic function of a flat kernel, which is a straightforward binary function:
\begin{align*}
K(x) = \begin{cases}
1 & \mbox{ if } \|x\| \leq \lambda \\
0 &\mbox{ if } \|x\| > \lambda.
\end{cases} 
\end{align*}

In two-dimensional feature space, non-parametric MSC only requires one parameter $h$ as a scalar of window size $\lambda$, which is called the kernel bandwidth. This parameter has clear physical interpretation in our case as the average width of the detected objects. To be adjustable in different scenarios, we tune it in a range of no less than the average width of one passenger vehicle, and no greater than the maximum length. A shortage of MSC is that it is not highly scalable, while it is guaranteed to converge. Thus a single trip should be cut into small segments in data processing.\par 

The main problem of one-step classification for parking detection is its inevitably high Type I Error. If clusters are directly assigned to the given parking spaces on the map, we implies that all the other spaces are unoccupied, which will cause unfavorable errors in the guidance. To avoid that problem, we add a second step finding maximum margin decision boundaries using multi-class soft margin Support Vector Machine (SVM). Decision boundaries between two neighboring classes are linear. Based on labels from step 1, we can find maximum margin between two neighboring clusters using pairwise one-vs-one analysis, which fits $(n_{\mbox{clusters}}-1)$ classifiers instead of $O(n_{\mbox{clusters}}^2)$ using one-vs-one analysis. The threshold of whether there are the free parking spaces between clusters is the ratio of distance between boundaries to average vehicle length or width.

\subsection{Training Data: Single-Vehicle Pilot Test} 

In order to calibrate a proposed detection algorithm's parameters, we conducted a pilot test using a remodeled vehicle from UMTRI's IVBSS project \cite{green2008integrated}. To build a training dataset, the positions of parked vehicles and the driving paths of the probe car are predetermined, which will provide ground-truth data when reconstructing the test. The probe vehicle for this experiment is equipped with six wide field-of-view ($80^{\circ}$), short-range ($\sim$$10 m$) radars (SRR), a standard vehicular GPS and a data acquisition system (DAS) connected to CAN bus with over 600 channels at 10 to 50 Hz. In total there are twelve trips for calibration use. Assuming that each sensor has the same accuracy and works independently, we can calibrate them accordingly. Distance from the probe car to parking spaces ($D$ in figure 1) is set to be 80 inches to simulate driving in a parking lot. The average speed is set to be 5 mph to 10 mph to collect sufficient data. A vehicle feature extraction classifier is incorporated in each SRR. A DAS with computer connections serves as a transmitter in a connected vehicle network. \par 

\begin{figure}[h]
\centering
 \includegraphics[width=3in]{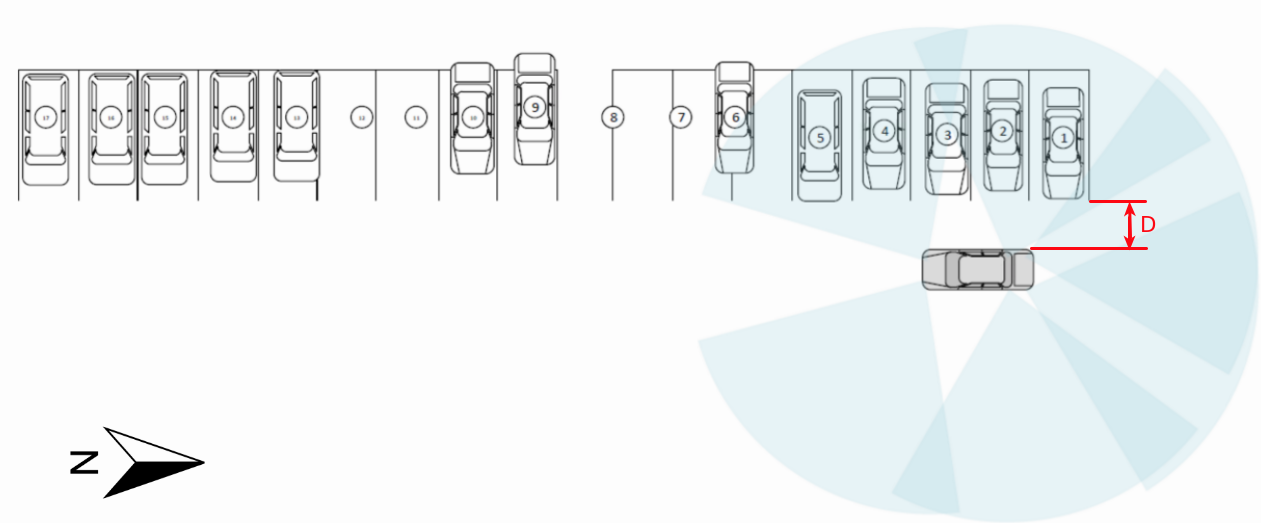}
\caption{Training data pilot test configuration; A probe vehicle equipped with 6 SRR (blue fans represent the field of view) scanning along 13 parked vehicle on 17 parking spaces.}
\end{figure}

\subsection{Test Data: Off-street and On-street Parking} 
After the model selection in training set, we want to inspect how this algorithm performs in real-world situations. Different from the pilot test, a three-day experiment is extended to a probe car driving in a parking lot (off-street parking) and on a two-way road with road-side parking (on-street parking) during peak hours. The path is predetermined as the two traces shown in figure 2. In total there are 160$\times$16  off-street parking spaces and 53$\times$6 on-street parking spaces valid observations in the test set.  The actual status of parking spaces are recorded by a camera and translated manually afterwards.  The result monitors 191 in/out events during the experiment, which guarantees to investigate the stability of the algorithm in handling different profiles. \par

\begin{figure}[!h]
\centering
	\subfloat[]{\includegraphics[width=1.65in]{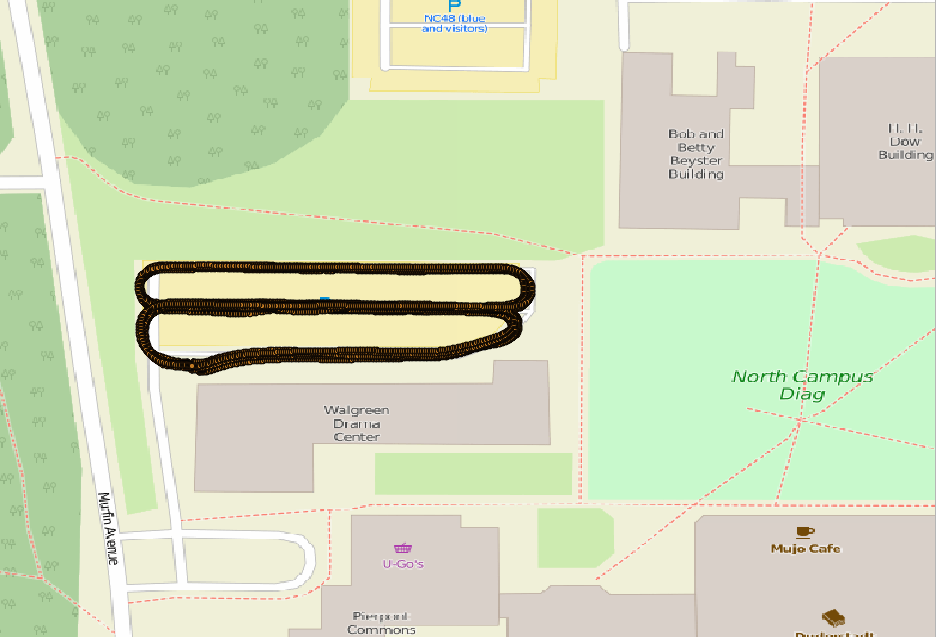}}
	\hspace{0.02in}
	\subfloat[]{\includegraphics[width=1.65in]{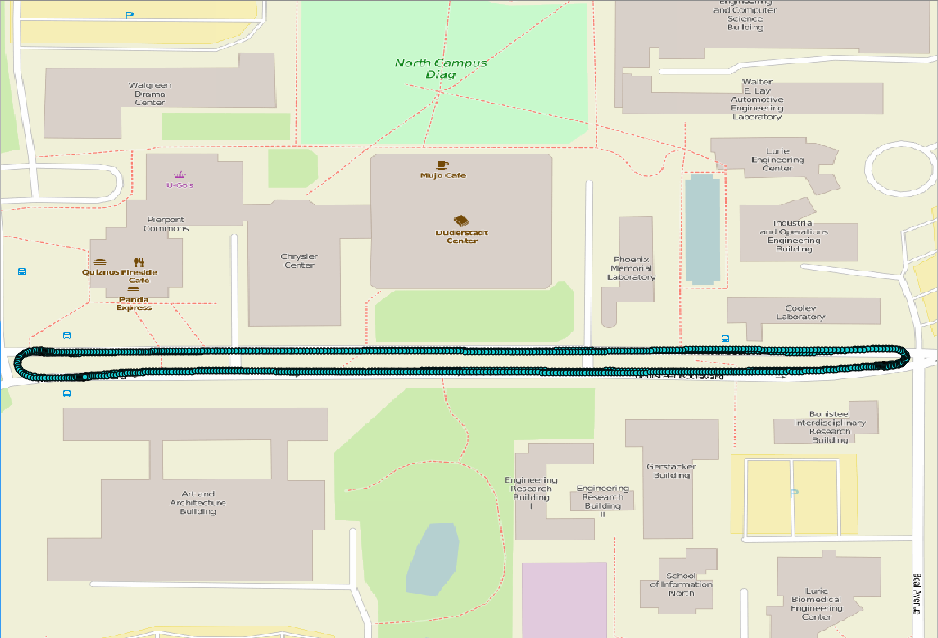}}
	\caption{GPS trace of off-street and on-street parking experiments. (a) NC-27 parking on North Campus, University of Michigan. (b) Bonisteel Blvd, Ann Arbor.}
	
\end{figure}

\subsection{Bayesian Updating} 

Individual probe cars exploration results are shared through a connected vehicle network, which can be treated as a cooperative mapping problem. Bayesian updating method can estimate the status of parking spaces recursively with new probe vehicle scanning data being accumulated.\par
Let $\p{m}$ represent an estimated map shared among probe cars, which is a node-edge map of parking spaces. $\p{s}_i^{(n)}$ denotes the state of probe car n at trip i, which consists of probe vehicles' paths (a list of nodes and directions) and origin-destination couples. $\p{z}_i^{(n)}$ represents the measurement model from vehicle n's radars. With regard to non-automated vehicle, mapping will not affect the control of the probe car itself so the state space is only the status of nodes on map.  \\
In sum, the Bayesian inference of updating parking space is:
\begin{align*}
p(\p{m}|\p{s}_{i}^{(n)}, \p{z}_{i}^{(n)}) = \frac{ p(\p{z}_{i}^{(n)} | \p{s}_{i}^{(n)}, \p{m})  p(\p{m}| \p{s}_{i}^{(n)} ) }{p(\p{z}_{i}^{(n)} | \p{s}_{i}^{(n)})}
\end{align*}

A distinct interpretation of mapping from traditional Bayesian state estimation is that the measurement term $p(\p{z}_{i}^{(n)} | \p{s}_{i}^{(n)}, \p{m} )$ in this case is the posterior at the end of each trip, and the prior $p(\p{m}| \p{s}_{i}^{(n)} )$ is the map being updated from trip 1 to trip $ i -1$.  Thus we have to assume that each probe car has the same measurement model, and furthermore assume that the probe cars will not interact with each others when driving.    

\subsection{Map Matching}
The map matching procedure in parking detection is straightforward because of the dimension-reduction in space. By assuming that the probe car is driving parallel to the row of parked car, we can apply one-to-one matching perpendicular to the direction of driving, and use 1/0 notation to represent the estimated state in each scanning. The detection range of right-side sensors is greater than left-side sensors, and from the comparison of errors, we find no significant difference in accuracy of detection if the object is within the range. Therefore, we don't need to consider the impact of $D$ in matching either.     

\section{Model Selection and Test Result}
Model selection including two parts: choosing the proper clustering method in step-one classification, and tuning the parameters of the classifier according to the training set. \par 

The criterion for choosing the best-performance clustering method is the correctness of matching the number of clusters and centroid locations with regard to the known parking configurations. The hypothesis testing for the result of detection algorithm is described in table I, and the Type I and Type II error rate are two indicators of the accuracy of the parking detection results. \par 
  
\begin{table}[h!]
\centering
\caption{Hypothesis Testing for Detection Algorithms}
\begin{tabular}{|l|c|c|c|}
\hline
\multicolumn{2}{|l|}{\multirow{2}{*}{}}                                & \multicolumn{2}{c|}{Null Hypothesis} \\ \cline{3-4} 
\multicolumn{2}{|l|}{}                                                  & occupied   &  unoccupied\\ \hline
\multicolumn{1}{|c|}{\multirow{3}{*}{Judgement}} & \begin{tabular}[c]{@{}c@{}} available\end{tabular} & Type I error & \begin{tabular}[c]{@{}c@{}}correct \\ inference\end{tabular} \\ \cline{2-4} 
\multicolumn{1}{|c|}{} & not available & \begin{tabular}[c]{@{}c@{}}correct \\ inference\end{tabular} & Type II error \\ \hline
\end{tabular}
\end{table} 

As a sample matching results shown in a sample trip in figure 3 and the entire set in figure 5, MSC outperforms others in the training set in both the states of occupancy and localization of parkings.   \par

\begin{figure}[h!]
\centering
\includegraphics[width=2.8in]{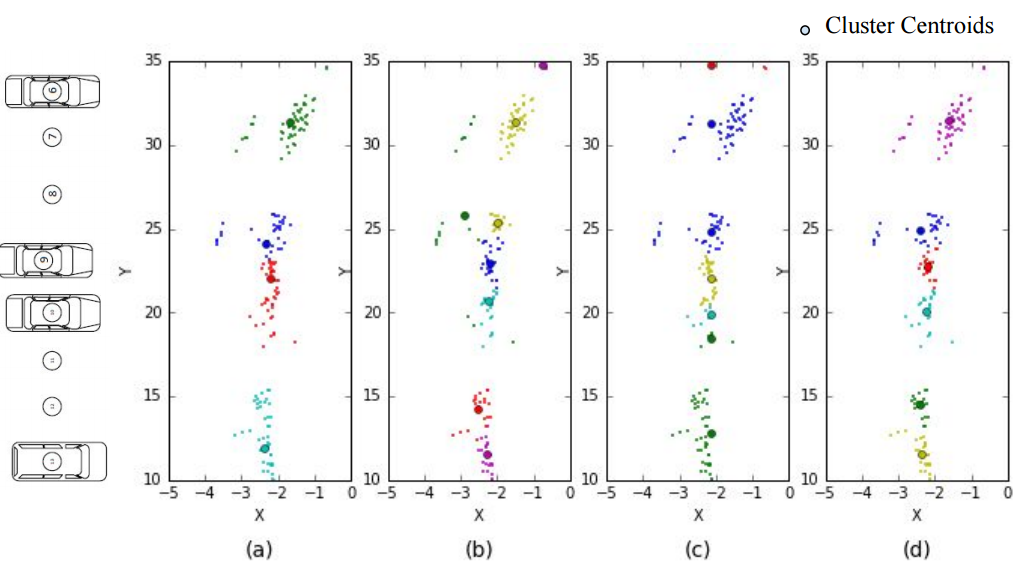}
\caption{Clustering results from a sample trip; From left to right, each color represent a different cluster: (a) MeanShift Clustering with flat kernel; (b) Gaussian Mixture Model with AIC; (c) Gaussian Mixture Model on Y-direction with AIC; (d) K-Means Clustering.}
\end{figure}

The Type I error rate using only MSC is about 10\% to 20\% in the training set, which is still unfavorable in terms of users' satisfactions. To avoid high Type I errors in inference, a second step SVM between neighboring clusters is applied and the final output reduces Type I error rate to less than 10\% in the training set when it successfully indicates the gap distance between clusters. The parameters tuning process is shown in figure 4. For off-street parking, the optimal bandwidth for MSC is set to be 2 $m$ because of facing the perpendicular sides of vehicles, and that parameter is 4.5 $m$ for on-street parking when facing the parallel sides of vehicles. We neglect the impact of other variables like driving speed when tuning the model.  \par

\begin{figure}[h!]
\centering
\vspace*{-0.2in}
\subfloat[]{\includegraphics[width=2.0 in]{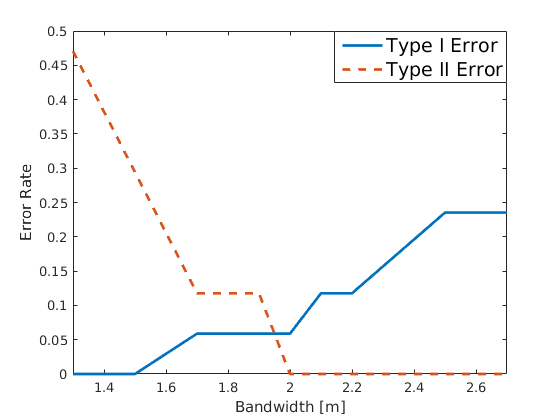}}
\subfloat[]{\includegraphics[width=1.0 in]{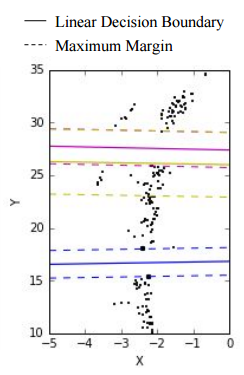}}
\caption{(a) Type I and Type II error rate over bandwidth in tuning $D$ in MSC. Bandwidth = 2 $m$ is optimal for off-street parking case. (b) SVM results of a sample trip.}
\end{figure}

\begin{figure}[!h]
	\vspace*{-0.1in}
	\subfloat[]{\includegraphics[width=1.8in]{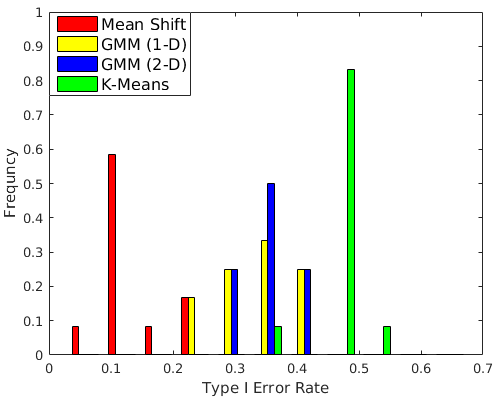}}
	\subfloat[]{\includegraphics[width=1.8in]{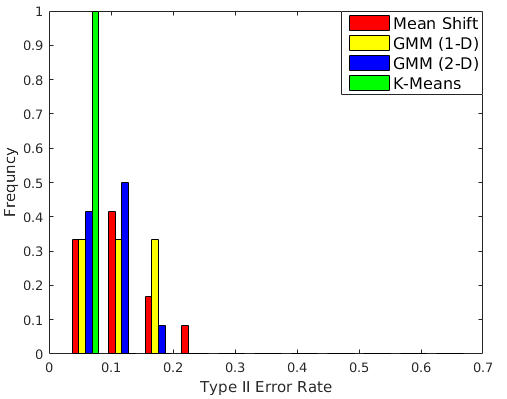}}
	\caption{(a) Type I error rate of the training data; (b) Type II error rate of training data.}
\end{figure}
 
The indicators of the performance of the proposed detection algorithm are Type I and Type II error rates of each segment in the off-street test set and on-street test set. Each trip is divided to segments when the heading of the probe car shifts over $\pm$90 degrees for the sake of avoiding errors at turning points. The histogram of the overall error rate on each segment are shown in figure 6 (on-street parking) and figure 7 (off-street parking). The results indicate that this statistical method's Type I error for on-street parkings is $32.62 \%$ in average, comparing to $15.23\%$ for the off-street cases. Type II error rates for on-street parking is almost 0, while higher for the off-street case. We have claimed that the Type I error is more concerned, so that the next step is to analyze the major parameters for errors. It is naturally to presume that the driving speed is dominating this difference in results while parameters of MSC have been tuned according to vehicle lengths, and a hypothesis testing is conducted on the speed profiles versus error rates.  \par

\begin{figure}[!h]
	\vspace*{-0.1in}
	\centering
	\includegraphics[width=2.5in]{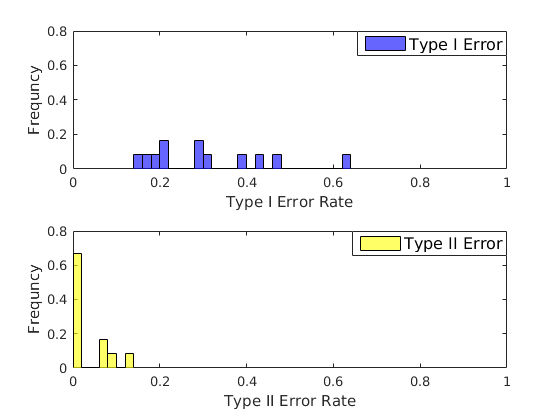}
	\caption{Matching results of on-street parking test data.}
\end{figure}

\begin{figure}[!h]
	\centering
	\includegraphics[width=2.5in]{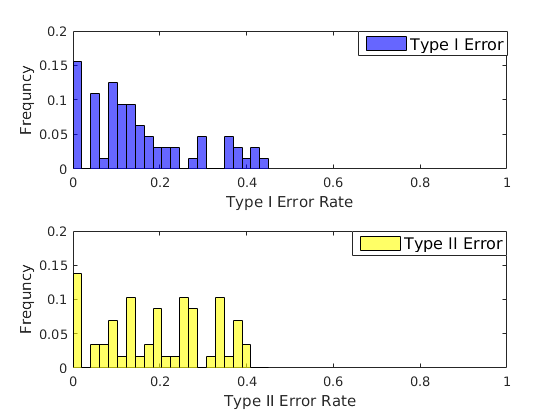}
	\caption{Matching results of off-street parking test data.}
\end{figure}

The regression analysis results within the groups  contradicts the assumption we made in figure 8 because the p-values for the first-order estimator is large. However, there is significant relationship between speed and detection errors as the p-value for the first-order estimator ($ =0.015 $) from linear regression is $2.65 e^{-5}$. It is not clear whether the error rate increases because of higher driving speed or different parameters in the classifiers for two cases.   \par
 
\begin{figure}[!h]
	\centering
	\includegraphics[width=2.8in]{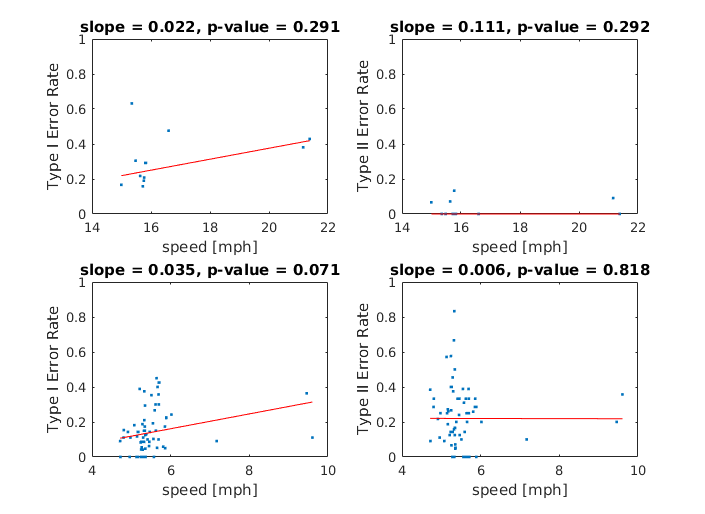}
	\caption{Detection error rate over speed regression on the on-street and off-street test set.}
\end{figure}

\section{CONCLUSIONS}
We have made these two major contributions to the application of sensor-based parking occupancy detection systems in this project:
\begin{enumerate}
\item Proposing a two-step classification method for sparse sensor data collected from a single probe car, which is proved to be more stable and effective in balancing Type I and Type II errors especially for off-street parkings. 
\item A comprehensive off-street / on-street parking occupancy detection system that utilizes the sensors in ADAS of multiple probe cars to form a collaborative parking detection network.
\end{enumerate}

The future work includes two potential directions: providing systematic management strategies to improve the estimations of parking in a long run, and building a back-end for broadcasting parking information. 

\addtolength{\textheight}{-12cm}   % This command serves to balance the column lengths

%%%%%%%%%%%%%%%%%%%%%%%%%%%%%%%%%%%%%%%%%%%%%%%%%%%%%%%%%%%%%%%%%%%%%%%%%%%%%%%%

%%%%%%%%%%%%%%%%%%%%%%%%%%%%%%%%%%%%%%%%%%%%%%%%%%%%%%%%%%%%%%%%%%%%%%%%%%%%%%%%

%%%%%%%%%%%%%%%%%%%%%%%%%%%%%%%%%%%%%%%%%%%%%%%%%%%%%%%%%%%%%%%%%%%%%%%%%%%%%%%%

\section*{ACKNOWLEDGMENT}
The authors would like to thank Mark Gilbert and his colleagues in Engineering Systems Group at UMTRI for assisting in setting up the experiments and all volunteer drivers in the field experiments.

%%%%%%%%%%%%%%%%%%%%%%%%%%%%%%%%%%%%%%%%%%%%%%%%%%%%%%%%%%%%%%%%%%%%%%%%%%%%%%%%

\bibliographystyle{./IEEEtran}
\bibliography{./IEEEabrv,./SRRpaper}

\end{document}